\documentclass[10pt,twocolumn,letterpaper]{article}

\usepackage{cvpr}
\usepackage{times}
\usepackage{graphicx}
\usepackage{amsmath}
\usepackage{amssymb}
\usepackage{bbding}
\usepackage{booktabs}
\usepackage{array}
\usepackage{comment}

\usepackage[breaklinks=true,bookmarks=false]{hyperref}

\cvprfinalcopy 

\def\cvprPaperID{***} 

\begin{document}

\title{RGB-based 3D Hand Pose Estimation\\via Privileged Learning with Depth Images}

\author{Shanxin Yuan\\
Imperial College London\\
{\tt\small shanxinyuan@gmail.com}
\and
Bj{\"o}rn Stenger\\
Rakuten Institute of Technology\\
{\tt\small bjorn@cantab.net}
\and 
Tae-Kyun Kim\\
Imperial College London\\
{\tt\small tk.kim@imperial.ac.uk}
}

\maketitle

\begin{abstract}

This paper proposes a method for hand pose estimation from RGB images that uses both external large-scale depth image datasets and paired depth and RGB images as privileged information at training time. 
We show that providing depth information during training  significantly improves performance of pose estimation from RGB images during testing.
We explore different ways of using this privileged information: (1) using depth  data to initially train a depth-based network, (2) using the features from the depth-based network of the paired depth images to constrain mid-level RGB network weights,  and (3) using the foreground mask, obtained from the depth data, to suppress the responses from the background area.
By using paired RGB and depth images, we are able to supervise the RGB-based network to learn middle layer features that mimic that of the corresponding depth-based network, which is trained on large-scale, accurately annotated depth data.
During testing, when only an RGB image is available, our method produces accurate 3D hand pose predictions. Our method is also tested on 2D hand pose estimation. 
Experiments on three public datasets show that the method outperforms the state-of-the-art methods for hand pose estimation using RGB image input.
\end{abstract}

\section{Introduction}

3D hand pose estimation has been greatly improving in the past few years, especially with the availability of depth cameras. 
While new methods~\cite{oberweger2015training,ye2016spatial,ge20173d,wan2017crossing,tang2018opening} and datasets~\cite{tang2014latent,tompson2014real,sun2015cascade,yuan2017bighand2,garcia2018first} have been published, state-of-the-art methods are still 
lacking in accuracy required for fine manipulations for AR or VR systems.
There is a large accuracy gap between pose estimation from RGB and depth image input, which several recent works have aimed to narrow~\cite{simon2017hand,zimmermann2017learning,mueller2017real,PantelerisArgyros2017}. 
One of the difficulties has been the lack of large-scale realistic RGB datasets with accurate annotations. Recent papers have addressed this issue by creating synthetic datasets~\cite{zimmermann2017learning}, or employing GANs to generate training data~\cite{mueller2017ganerated}.
In this paper we propose using depth data as {\em privileged information} during training.
Fully annotated depth datasets~\cite{tang2014latent,tompson2014real,sun2015cascade,yuan2017bighand2,garcia2018first} are abundant in the literature, but so far no attempt has been made to use this data to support the task of 3D hand pose estimation from RGB images. 
There are also a few RGB-D datasets proposed recently~\cite{zimmermann2017learning,zhang20163d} to tackle the problem of 3D hand pose estimation from RGB images, however all existing methods~\cite{zimmermann2017learning,mueller2017ganerated,zhang20163d} utilise only RGB images for training. The available depth images, either paired with RGB images~\cite{zhang20163d,zimmermann2017learning} or alone in the large-scale \textit{BigHand2.2M} dataset~\cite{yuan2017bighand2} could be used to aid the training. 

\begin{figure*}[t]
\centering
        \includegraphics[trim=1.0cm 5.5cm 5.5cm 2cm, clip=true,width=1.0 \textwidth]{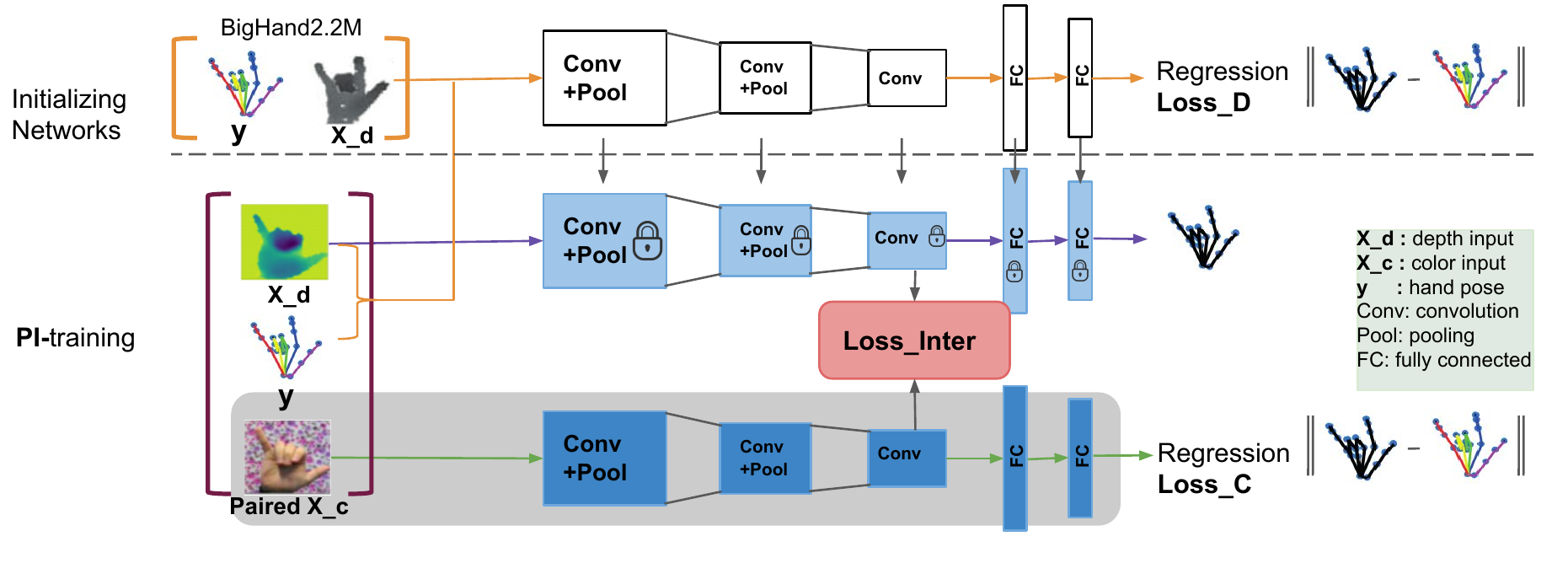}

  \caption{
  \textbf{Proposed framework for 3D hand pose estimation from an RGB image using privileged depth data.} \textit{Training proceeds in two stages, a pre-training stage and privileged information (PI)-training stage. In the first stage, a depth-based network (top) and an RGB-based (bottom) network are trained independently to minimize 3D pose loss \textit{Loss\_D} and \textit{Loss\_C}. In the second stage, we freeze the parameters of the depth-based network and continue training with paired RGB and depth images, by minimizing a joint loss, which includes \textit{Loss\_C}  and a mid-level feature regression loss \textit{Loss\_Inter}.}}
    \label{fig:PI_cnn_model}
\end{figure*}

The use of privileged information in training~\cite{vapnik2009new}, also called training with hidden information~\cite{wang2015classifier}, or side information~\cite{xu2013speedup}, has been shown to improve performance in other domains, such as image classification~\cite{chen2017training}, object detection~\cite{hoffman2016learning}, and action recognition~\cite{shi2017learning}. 
But the concept of using privileged information to help 3D hand pose estimation from RGB images has not been attempted. 
To the best of our knowledge, this paper proposes the first solution. Existing methods for 3D hand pose estimation from RGB images pursue two main directions: (1) using only RGB images for 3D hand pose estimation~\cite{zhang20163d,zimmermann2017learning,mueller2017ganerated}, with different CNN models being proposed. Given the limited size of real RGB datasets, a large number of synthetic images~\cite{zimmermann2017learning,mueller2017ganerated} are created to help the training, whether they are purely synthetic~\cite{zimmermann2017learning}, or using CycleGAN~\cite{zhu2017unpaired} to enforce a certain realism ~\cite{mueller2017ganerated}. (2) Using RGB-D images for  3D hand pose tracking~\cite{mueller2017real}, where the input is the depth channel in addition to the RGB channels. This works well when the paired RGB and depth images are available at test time. The lack of large-scale annotated training data limits the success of this approach. Our study proposes a new framework for 3D hand pose estimation from RGB images, by using the existing abundant  fully annotated depth data in training, as privileged information. This helps improve 3D hand pose estimation using a single RGB image input at test time.
Our method transfers supervision from depth images to RGB images. We use two networks, an RGB-based network and a depth-based network, see Figure~\ref{fig:PI_cnn_model}. 
We explore different ways to use depth data: (1) initially, we treat a large amount of independent external depth training data as privileged information to train the depth-based network. (2) After the initial training is completed, paired RGB and depth images are used to tune the RGB-based network and the depth-based network. The idea is to let the middle layer activations of the RGB network mimic that of the depth network. (3) We also explore the use of foreground hand masks to suppress background area activations in the middle layers of the RGB network. By doing this, we force the RGB network to extract features only from the foreground area.

Compared to existing methods for 3D hand pose estimation by RGB images, our main contributions are:
\begin{itemize}
  \item To the best of our knowledge, this paper is the first to introduce the concept of using privileged information (depth images) to help the training of a RGB-based hand pose estimator.   
  \item We propose three ways to use the privileged information: as external training data for a depth-based network, as paired depth data to transfer supervision from the depth-based network to the RGB-based network, as hand masks to suppress the background activations in the RGB-based network.
  \item Our training strategy can be easily embedded into existing pose estimation methods. We demonstrate this in the experiments of 2D hand pose estimation with an RGB image input by a different CNN model. Results on 2D hand pose estimation, using our training strategy improve over state-of-the-art methods for 2D hand pose estimation with RGB input.  
\end{itemize}

Comprehensive experiments are conducted on three datasets: the Stereo dataset~\cite{zhang20163d}, the RHD dataset~\cite{zimmermann2017learning}, and the Dexter-Object dataset~\cite{sridhar2016real}. The Stereo dataset and RHD dataset are used for evaluating 3D pose estimation from an RGB input. All three datasets are used for evaluating 2D hand pose estimation from a single RGB image. 

\section{Related Work}

\textbf{3D hand pose estimation.} 
Hand pose estimation from depth data has made rapid progress in the past years~\cite{oberweger2015training,ge20173d,wan2017crossing,sharp2015accurate,choi2015collaborative}, where comprehensive studies~\cite{erol2007vision,supancic2015depth,yuan20183d}    have been instrumental in advancing the field.
Random forests~\cite{tang2014latent,tang2015opening,wan2016hand} and CNNs~\cite{ye2016spatial,ge20173d,wan2017crossing,tompson2014real} trained on large-scale public depth image
datasets~\cite{tang2014latent,tompson2014real,sun2015cascade,yuan2017bighand2,garcia2018first} have shown good performance.
A recent benchmark evaluation~\cite{yuan20183d} showed that modern methods achieve mean 3D joint position errors of less than 10mm.
Hand pose estimation from RGB images is significantly more challenging
~\cite{simon2017hand,zimmermann2017learning,mueller2017real,PantelerisArgyros2017}. 
Due to the difficulty in capturing real RGB datasets with accurate 3D annotations, recent methods employ synthetic CG data~\cite{zimmermann2017learning}, or \textit{GANerated} images~\cite{mueller2017ganerated}, which are more realistic synthetic images created with a CycleGAN~\cite{zhu2017unpaired}.
Mueller \etal~\cite{mueller2017ganerated} use an image-to-image translation network to create a large amount of RGB training images and combine a CNN with a kinematic 3D hand model for pose estimation. The method requires a predefined hand model, adapted for each user.
Simon \etal's {\textit{OpenPose}}~\cite{simon2017hand} system generates an annotated RGB dataset using a panoptic studio setup, using multiple views to bootstrap 2D hand pose estimation. 
Zimmermann and Brox~\cite{zimmermann2017learning} proposed combining hand segmentation and 2D hand pose estimation (using \textit{CPM}~\cite{wei2016convolutional}), followed by estimating 3D hand pose relative to a canonical pose. 
Panteleris and Argyros~\cite{PantelerisArgyros2017} estimate absolute 3D hand pose by first estimating 2D hand pose and then optimizing a 3D hand model with inverse kinematics. 
Note that there also exists a large body of work on the related task of recovering full 3D human body pose from images.
One line of work aims to directly estimate the 3D pose from images~\cite{li20143d,zhou2016deep,toshev2014deeppose}.
A second approach is to first estimate 2D pose, often in terms of joint locations, and then lift this to 3D pose.
2D key points can be reliably estimated using CNNs and 3D pose is estimated using structured learning or a kinematic model~\cite{tome2017lifting,tompson2014joint,simo2012single,zhou2016sparseness}.\\

\textbf{Learning with privileged information and transfer learning.}
Privileged information denotes training data that is available only during training but not at test time.
The concept to provide teacher-like supervision at training time was introduced by Vapnik and Vashist~\cite{vapnik2009new}.
The idea has proven useful in other domains~\cite{chen2017training,hoffman2016learning,shi2017learning}.
Shi \etal~\cite{shi2017learning} treated skeleton data as privileged information in CNN-RNNs for action recognition from depth sequences. Chen \etal~\cite{chen2017training} manually annotated object masks in 10\% of the training data and treated these as privileged information for image classification. 
The idea is related to network compression and mimic learning proposed by Ba and Caruana~\cite{ba2014nips} as well as network distillation by Hinton \etal~\cite{hinton2015distillation}, where intermediate layer outputs of one network are approximated by another, possibly smaller, network.
These techniques can be used to significantly reduce the number of model parameters without a significant drop in accuracy.
In our case, the application target is similar to transfer learning and domain adaptation. Information from one task, prediction from depth images, is shared with another, prediction from RGB images.
In transfer learning and domain adaptation information is shared across different data modalities~\cite{rad2018domain,chen2014recognizing,hoffman2016learning}. Chen \etal~\cite{chen2014recognizing} proposed recognition in RGB images by learning from labeled RGB-D data. A common feature representation is learned across two feature modalities.
Hoffman \etal~\cite{hoffman2016learning} learned an additional \textit{hallucination} representation, which is informed by the depth data in training. At testing, it used the softmax to select the final prediction between the predictions from the hallucination representation and the predictions from RGB representation.
Luo \etal~\cite{luo2017graph} recently proposed graph distillation for action detection with privileged modalities (RGB, depth, skeleton, and flow), where a novel graph distillation layer was used to dynamically learn to distill knowledge from the most effective modality, depending on the type of action.
In our case, we use paired depth and RGB images during training. Depth and RGB networks are first trained separately. Subsequently the RGB network are progressively updated, while the depth network parameters remain fixed.

\textbf{Learning a latent space representation.} 
Latent space representation also shows promising for 3D hand pose estimation from RGB images~\cite{spurr2018cross,iqbal2018hand}.
Spurr \etal~\cite{spurr2018cross} learned a cross-modal statistical hand model, via learning of a latent space representation that embeds sample points from multiple data sources such as 2D keypoints, images, and 3D hand poses. Multiple encoders were used to project different data modalities into a unified low-dimensional latent space, where a set of decoders reconstruct the hand configuration in either modality.
Iqbal \etal~\cite{iqbal2018hand} used latent 2.5D heatmaps, containing the latent 2D heatmaps and latent depth maps, to ensure the scale and translation invariance. Absolute 3D hand poses are reconstructed from the latent 2.5D heatmaps.
Cai \etal~\cite{Caiweakly} proposed a weakly-supervised method for 3D hand pose estimation from RGB image by introducing an additional depth regularizer module, which rendered a depth image from the estimated 3D hand pose. Training was conducted by minimizing an additional loss term, which is the $L1$ distance between the rendered depth image and the ground truth depth image.

\section{Methods}

We propose a framework to train a hand pose estimation model from RGB images by using depth images as privileged information. 
The model learns a new RGB representation which is influenced by the paired depth representation through mimicking the mid-level features of a depth network. 

As shown in Figure~\ref{fig:PI_cnn_model}, we use depth images in two ways: (1) to  train an initial depth-based network with the aim of regressing 3D hand poses. Depth data that is annotated with 3D full hand pose information is abundant in the literature, and we choose the largest real dataset BigHand2.2M~\cite{yuan2017bighand2} to train our depth-based model, see the top row of Figure~\ref{fig:PI_cnn_model}. (2) Paired RGB and depth images are fed into the RGB-based and depth-based network with the parameters of the depth-based network being frozen. The training of the RGB-based network continues with the aim of minimizing a joint loss function. The joint loss function has two parts, the first part being the 3D hand pose regression loss, \textit{Loss\_C}, and the second part the mid-level regression loss, \textit{Loss\_Inter}.

\subsection{Architecture}

Figure~\ref{fig:PI_cnn_model} shows our training architecture. There are two base models, each for one input channel. We use deep convolutional neural networks (CNNs), which have been widely used in hand pose estimation and have proven useful in transferring information from one network to another~\cite{hinton2015distillation}. 
Prior work~\cite{mueller2017real} has been shown useful in combining RGB and depth images as a four-dimensional RGB-D input to a single CNN model to estimate 3D hand pose. 
In our architecture, we share information in the middle layers of our two CNN models, one is a depth-based network and the other one is an RGB-based network. Each CNN model takes an input (a depth image or an RGB image) and produces a 3D hand pose estimation result.

For clarity, we denote the depth-based network \textit{Depth\_Net}, the RGB-based network \textit{RGB\_Net} when this is trained before privileged information is used. When privileged information is introduced in the training, we denote the RGB-based network \textit{RGB\_PI\_Net}. In summary, \textit{RGB\_Net} and \textit{RGB\_PI\_Net} are the same CNN model trained before and after the paired RGB and images are used to train the RGB channel.

We aim at sharing information between the middle layers of our two CNN models, and in particular using \textit{Depth\_Net} to inform \textit{RGB\_PI\_Net} in the training time when paired RGB and depth images are available. 
To let the \textit{Depth\_Net} channel share information with \textit{RGB\_PI\_Net}, we introduce an intermediate regression loss between the paired layers in the two models. 
This intermediate regression loss is inspired by prior works~\cite{hoffman2016learning,hinton2015distillation}, where similar techniques are used for model distillation~\cite{hinton2015distillation}, supervision transfer from well labeled RGB images to depth images with limited annotation~\cite{gupta2016cross}, and hallucination of different modalities~\cite{hoffman2016learning}.
We therefore introduce an intermediate loss,
which helps \textit{RGB\_PI\_Net} to extract middle level features that mimic the responses of the corresponding layer of the \textit{Depth\_Net} using the paired depth image.

The intermediate loss (or \textit{Loss\_Inter} as shown in Figure~\ref{fig:PI_cnn_model}) is defined as:
\begin{equation}
Loss\_Inter(k)= \|A_{k}^{Depth} - A_{k}^{RGB}\|_{2}^{2} \,,
\label{equ:interloss}
\end{equation}
where $A_{k}^{Depth}$ and $A_{k}^{RGB}$ are the $k_{th}$ layer activations for Depth Network and RGB Network, respectively. 
During testing, where only an RGB image is available, we feed the RGB image into \textit{RGB\_PI\_Net} to estimate the 3D hand pose.

\subsection{Training with privileged information}

This section explains the details of training the proposed architecture. 
We choose a base CNN for \textit{Depth\_Net} and \textit{RGB\_PI\_Net} for 3D hand pose estimation. For the base model, we build on Convolutional Pose Machine (CPM)'s~\cite{wei2016convolutional} feature extraction layers with two fully connected layers to regress a 63 dimensional 3D hand pose with 21 joints.

In this initial stage, we call this external depth images as privileged information. Our \textit{Depth\_Net} is initially independently trained on BigHand2.2M~\cite{yuan2017bighand2} dataset, which has 2.2 million fully annotated (21 joints) depth images. 
After training, the model is further trained on the depth images of a smaller dataset (\eg, Stereo~\cite{zhang20163d} and RHD~\cite{zimmermann2017learning} datasets) that has fully annotated paired RGB and depth images.
The \textit{RGB\_Net} is initially trained on the RGB images from the same dataset.

When the initial training is completed for both CNN models, we freeze the parameters of the \textit{Depth\_Net} and start training \textit{RGB\_PI\_Net} with privileged information. 
In this stage, our privileged information is the paired depth images, and comes into use in the form of the middle layer activations of the \textit{Depth\_Net}. 
During the privileged training stage, we want the \textit{RGB\_PI\_Net}'s middle level layer's activations to match the activations of the corresponding layers of the \textit{Depth\_Net}. 
We have two losses to optimize: (1) \textit{Loss\_Inter} (Eqn.~\ref{equ:interloss}) is used to match the middle layer activations of the two CNN models. (2) \textit{Loss\_C} (see Figure~\ref{fig:PI_cnn_model}) is the $L2$ loss between the ground truth and the estimated 3D hand pose. Here we use a joint loss:
\begin{equation}
Loss\_Joint(k)= Loss\_Inter(k) + \lambda \cdot Loss\_C,
\label{equ:jointloss}
\end{equation}
where $\lambda$ is used to balance the two losses, a larger value of $\lambda$ means less supervision is required from the privileged information, a smaller value means that the model depends more on the supervision. We set $\lambda$ to 100 for all experiments.

\begin{figure*}[t]
\centering
        \includegraphics[trim=3cm 5.5cm 5.5cm 4.3cm, clip=true,width=.9\textwidth]{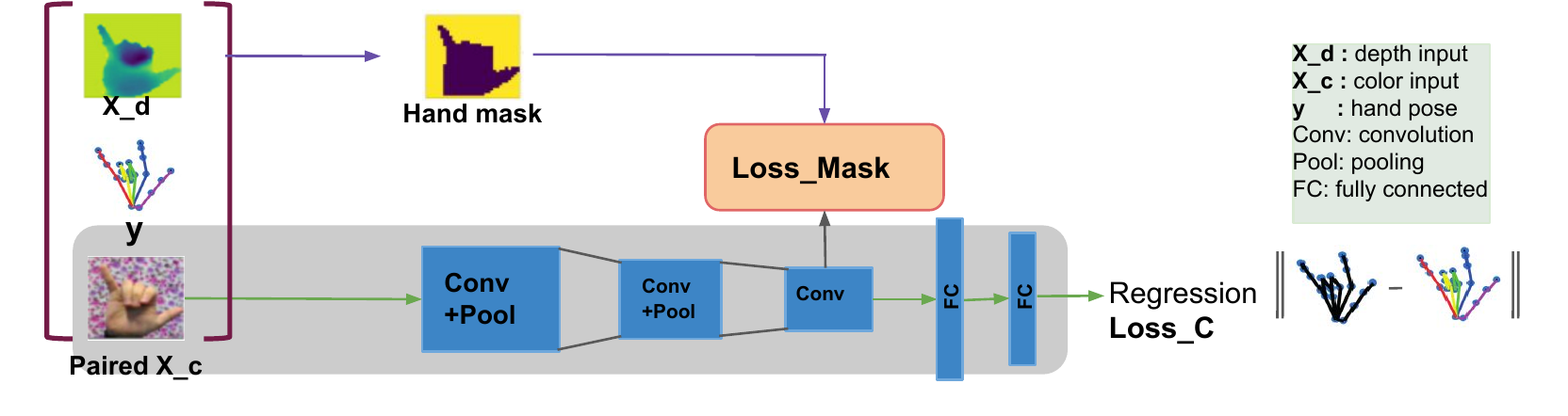}

  \caption{\textbf{Treating hand mask as privileged information}. Hand mask are used as privileged information to suppress the responses from the background area in the middle layers.}
  \label{fig:PI_cnn_sec_feature}
\end{figure*}

\subsection{Foreground mask as privileged information}
 
In addition to the supervision from depth images, we also explore the idea of extracting hand masks from depth images and embedding the hand masks into CNN layers of \textit{RGB\_PI\_Net} to suppress the background features. 
As shown in Figure~\ref{fig:PI_cnn_sec_feature}, we treat the hand mask $M_{h}$ as privileged information. 
At test time, when the hand mask is not available, the CNN model is viewed as a standard CNN with convolutional layers, pooling layers and full-connected layers, where the \textit{Loss\_Mask} is not used.
In the training stage, the foreground hand mask is introduced in the last convolutional layer, as shown in Figure~\ref{fig:PI_cnn_sec_feature}. Pixels of the mask $M_{h}$ are zero on the hand region, and one otherwise. We suppress background features by minimizing the regression loss \textit{Loss\_Mask}:
\begin{equation}
Loss\_Mask= \|A_{k}^{RGB} \odot M_{h}\|_{2}^{2} \\ \; ,
\label{equ:maskloss}
\end{equation}
where $\odot$ denotes element-wise multiplication.

By minimizing the regression loss, where the response on the hand is multiplied by zero and the response outside the hand is multiplied by one,  the response from outside the hand area is suppressed, focusing the response on the hand region. 

\begin{table}[t]
  \centering
  \small
  \resizebox{\columnwidth}{!}{
  \begin{tabular}{lrrrclc}
  \toprule 
  \bf  Dataset & \bf  No. Training & \bf  No. Test  &  \bf  No. Joints & \bf  Annotation & \bf  Type \\
  \midrule
    Stereo~\cite{zhang20163d} & 15,000 & 3,000  & 21 & 2D, 3D & real \\  
    RHD~\cite{zimmermann2017learning} & 41,258 & 2,728  & 21 & 2D, 3D & synthetic \\  
  Dexter-Object~\cite{sridhar2016real} & - & 3,111 & 5 (tips) & 2D, 3D & real\\
   \bottomrule  
   \end{tabular}}
    \caption{\textbf{Public datasets used in our experiments.}  }
  \label{tab:datasets}
\vspace{-5mm}
\end{table}

\section{Experiments}

We carried out experiments on both 3D and 2D hand pose estimation from RGB images.
Our experiments are conducted on three public RGB-D datasets: the RHD dataset~\cite{zimmermann2017learning}, the Stereo dataset~\cite{zhang20163d}, and the Dexter-Object dataset~\cite{sridhar2016real}, as shown in Table~\ref{tab:datasets}.
The RHD dataset is created synthetically and contains 41,238 training and 2,728 test images, with a resolution of $320 \times 320$. Each pair of RGB and depth images contains 3D annotations for 21 hand joints, and intrinsic camera parameters. The RHD dataset is built from 20 different subjects performing 39 actions. The training set has 16 subjects performing 31 actions, while the test set has 4 subjects performing 8 actions. The dataset contains diverse backgrounds sampled from 1,231 Flickr images.
The Stereo~\cite{zhang20163d} dataset is a real RGB-D dataset, which has 18,000 pairs of RGB and depth images with a resolution of $640 \times 480$ pixels. Each pair is fully annotated with 21 joints. The dataset contains six different backgrounds with respect to different difficulties (\eg textured/textureless, dynamic/static, near/far, hightlights/no-highlights). For each background, there are two sequences, each containing 1,500 image pairs. The dataset is manually annotated. In our experiments, we follow the evaluation protocol of~\cite{zimmermann2017learning}, \ie, we train on 10 sequences (15,000 images) and test on the remaining 2 sequences (3,000 images).
The Dexter-Object~\cite{sridhar2016real} dataset contains 3,111 images of two subjects performing manipulations with a cuboid. The dataset provides RGB and depth images, but only fingertips are annotated. The RGB images have a resolution of $640\times320$ pixels. Due to the incomplete hand annotation, we use this dataset for cross-dataset generalization. 

During testing on a GTX 1080 Ti, the network forward steps take 6ms for 3D pose estimation and 8ms for the 2D case. The image cropping and normalization is the same as in~\cite{zimmermann2017learning}. To crop the hand region, we use ground truth annotations to obtain an axis aligned crop, resized to 256$\times$256 pixels by bilinear interpolation. Examples are shown in the first row of Figure~\ref{fig:feature_activations}. 
For 3D hand pose estimation, we use the root joint's world coordinates and the hand's scale to normalize the results.

\subsection{3D hand pose estimation from RGB}

In this section, we investigate the usefulness of depth images to improve the performance of 3D hand pose estimation from an RGB image.
Our base CNN model is built upon the feature extraction layers of Convolutional Pose Machine (CPM)~\cite{wei2016convolutional} with two fully connected layers. The final output is a 63 dimensioal vector denoting the 21 joint 3D locations. Specifically, our base CNN model contains 14 convolutional layers, 4 pooling layers, and 2 fully-connected layers. 
At training stage, we have access to paired RGB and depth images.
Initially the \textit{Depth\_Net} is trained on \textit{BigHand2.2M}~\cite{yuan2017bighand2}.
We continue to train the \textit{Depth\_Net} using the depth images from the small dataset, \eg, Stereo dataset or RHD dataset. We train the \textit{RGB\_Net} with the RGB images from the small dataset.
When the initial training is completed, we start PI-training with the paired RGB and depth images. We freeze the weights of the \textit{Depth\_Net} and add the intermediate regression loss \textit{Loss\_Inter} among the mid-level features of \textit{Depth\_Net} and \textit{RGB\_PI\_Net}, then we continue the training of \textit{RGB\_PI\_Net} by minimizing the joint loss $Loss\_Joint$. We apply the intermediate loss to the last convolutional layers of both branches, where the parameter $k$ is set to 18 in Equation~\ref{equ:interloss} and Equation~\ref{equ:jointloss}.

\begin{figure*}[t]
 \includegraphics[trim=2.9cm 2cm 5.5cm 4cm, clip=true, width=0.32\textwidth]{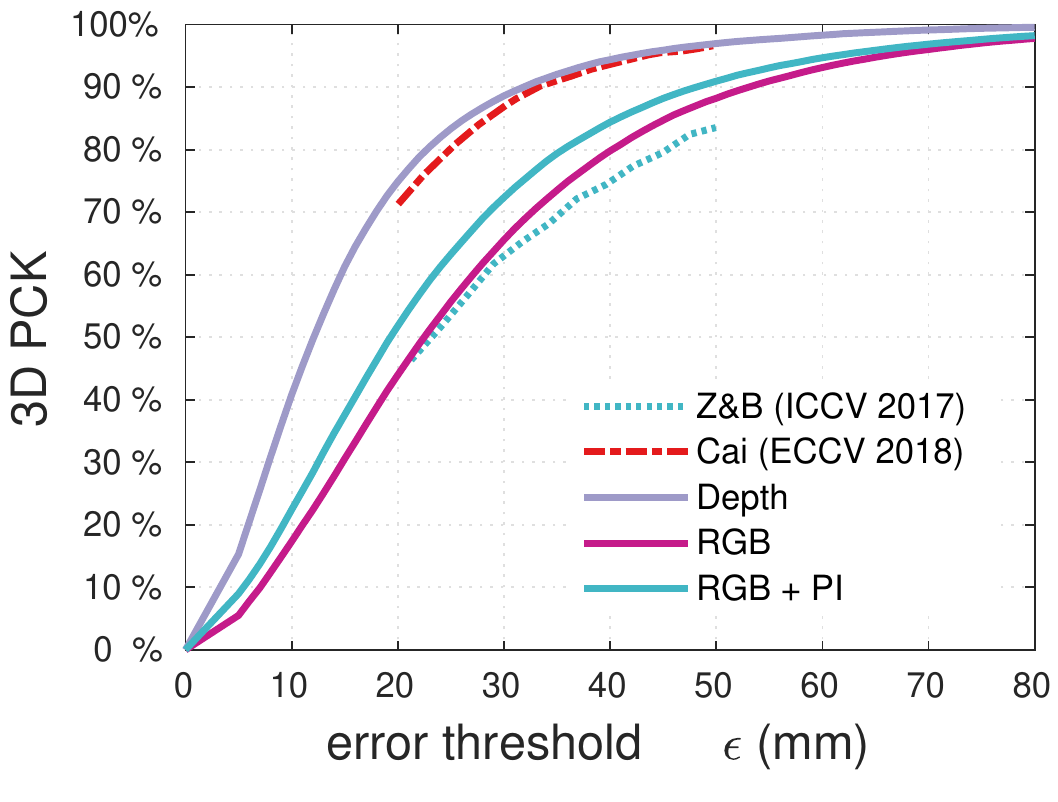}
\includegraphics[trim=2.9cm 2cm 5.5cm 4cm, clip=true, width=0.32\textwidth]{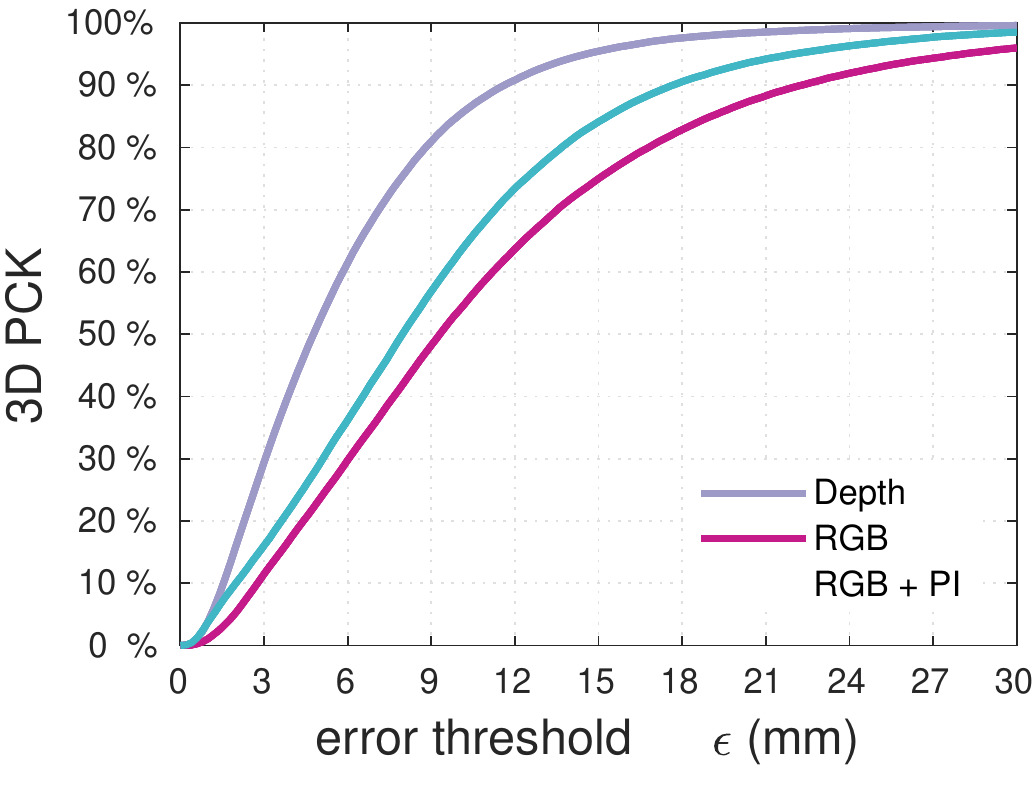}
\includegraphics[trim=2.9cm 2cm 5.5cm 4cm, clip=true, width=0.32\textwidth]
{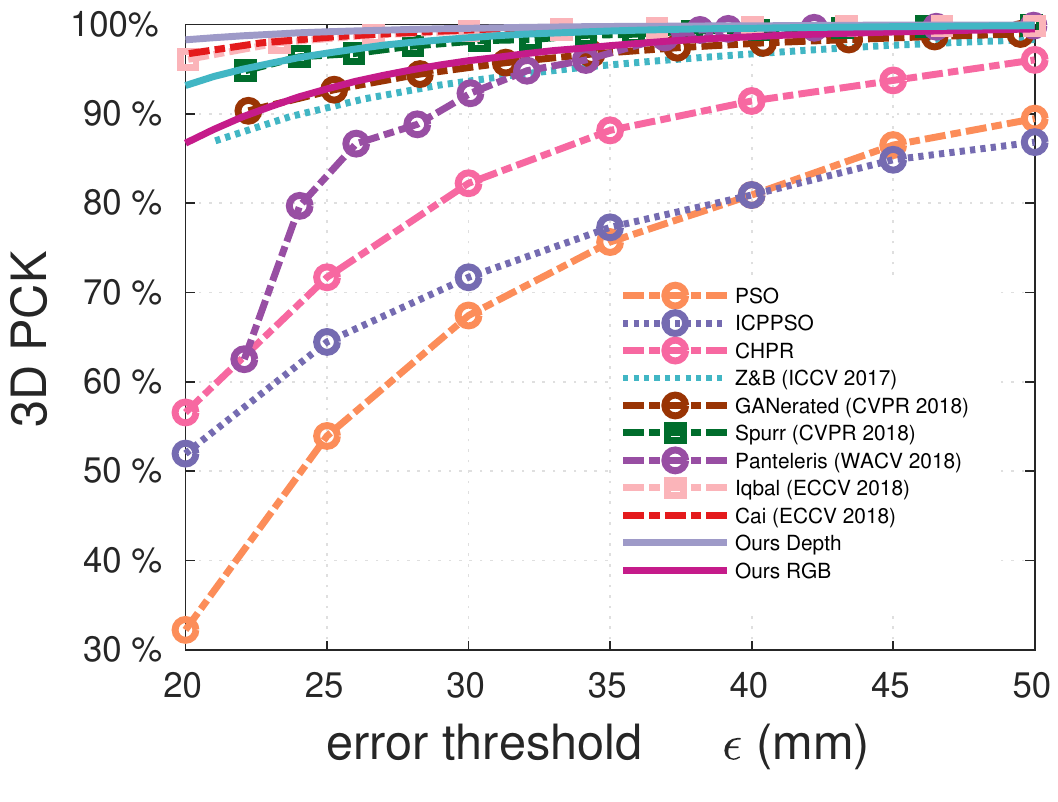}
\hfill

\includegraphics[trim=2.9cm 2cm 5.5cm 4cm, clip=true, width=0.32\textwidth]{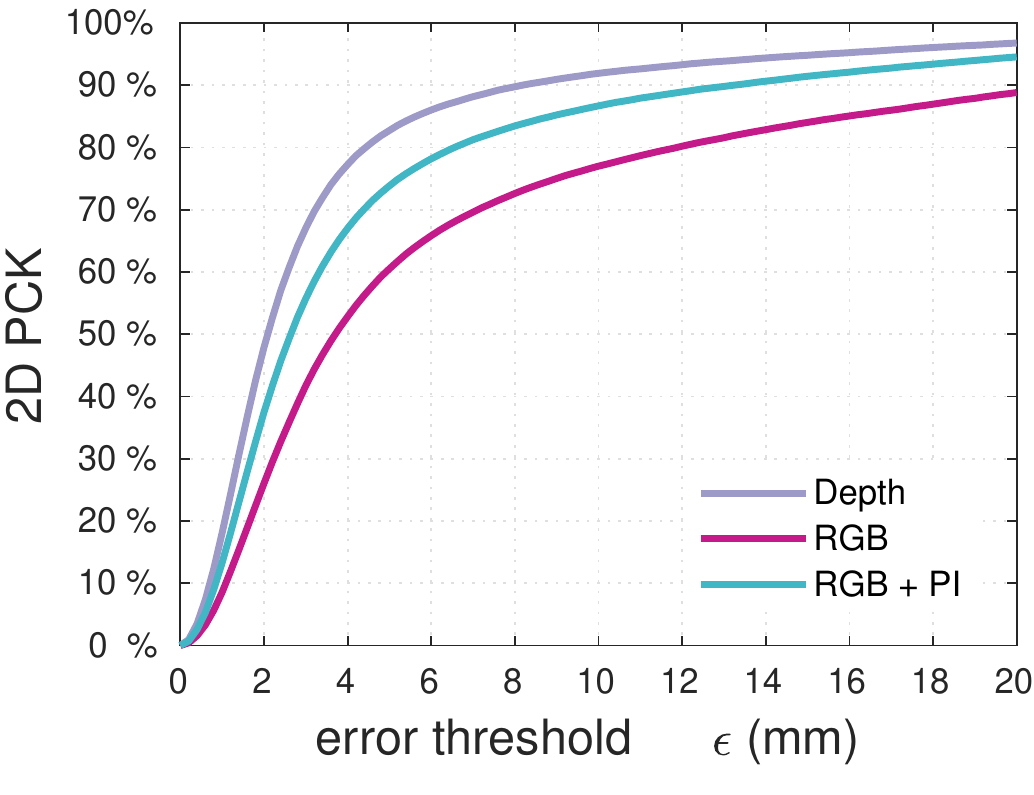}
\includegraphics[trim=2.9cm 2cm 5.5cm 4cm, clip=true, width=0.32\textwidth]{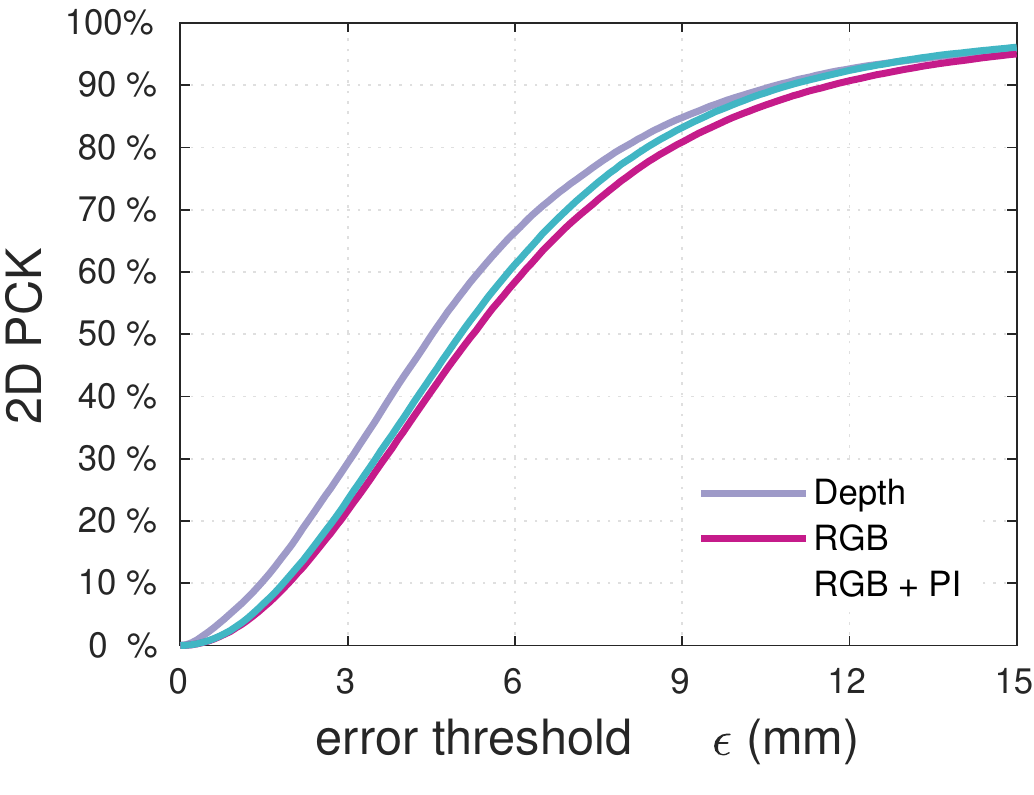}
\includegraphics[trim=2.9cm 2cm 5.5cm 4cm, clip=true, width=0.32\textwidth]{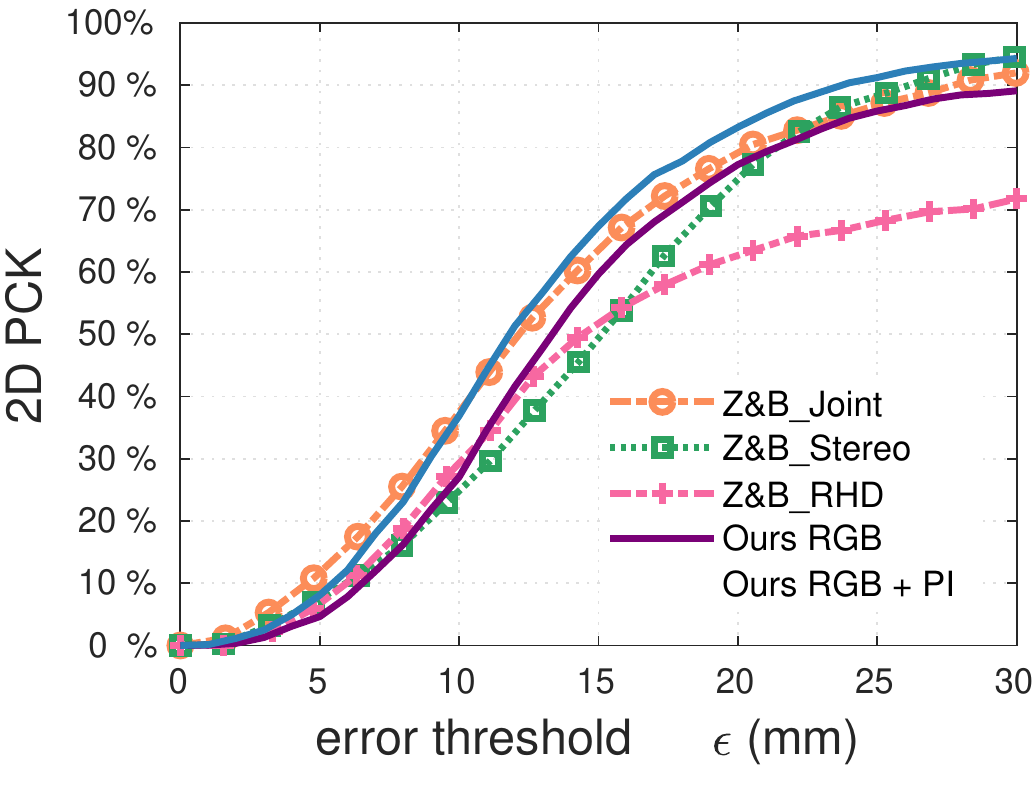}
   \hfill
  \caption{\textbf{Results one Stereo and RHD dataset for 3D hand pose and 2D pose accuracy}.\textit{Top row shows the comparisons of 3D hand pose accuracy, bottom row shows the comparisons of 2D hand pose accuracy. Top-left is self-comparison on RHD dataset, top-middle is self-comparison on Stereo dataset, top-right is comparison with state-of-the-art on Stereo dataset. Bottom-left is self comparison on RHD dataset, bottom-middle is self comparison on Stereo dataset, bottom-right is comparison with state-of-the-art on the Dexter-Object dataset.}}
  \label{fig:sota_com}
\end{figure*}

\textbf{Effect of PI-Learning:}
We conduct experiments with the two baseline CNNs and the CNN after PI training, see the accuracy curves in Figure~\ref{fig:sota_com} (top-left plot). 
Our networks only estimate relative 3D pose from a cropped RGB image patch containing the hand, to yield 3D hand pose in  world coordinates, we follow a similar procedure of~\cite{zimmermann2017learning}, \ie, by adding the absolute position of the root joint to our estimated results.
For comparison we choose the Percentage of Correct Keypoints (PCK) over a varying threshold.
Training with depth data significantly improves the performance of the RGB-based network, narrowing the gap to the depth-based network.  

\textbf{Comparison with the state of the art:}
We compare our results with state-of-the-art methods, including PSO~\cite{oikonomidis2011efficient}, ICPPSO~\cite{qianrealtime}, Zhang \etal~\cite{zhang20163d},  Z\&B~\cite{zimmermann2017learning}, GANerated~\cite{mueller2017ganerated}, Cai \etal~\cite{Caiweakly}, Spurr \etal~\cite{spurr2018cross}, Iqbal \etal~\cite{iqbal2018hand}, Panteleris \etal~\cite{Panteleris2018},  see Figure~\ref{fig:sota_com} (top-right plot). 
Our method out-performs all existing state-of-the-art methods. We outperform (Z\&B)~\cite{zimmermann2017learning} and~\cite{mueller2017ganerated}. While both~\cite{zimmermann2017learning} and~\cite{mueller2017ganerated} used extra training data, \cite{zimmermann2017learning} used both Stereo (real) and RHD (synthetic) data to train their network. \cite{mueller2017ganerated} used synthetic (GANerated) data to train their network. The proposed method uses less RGB training data and achieved the best performance. we significantly outperformed both methods with our privileged training strategy.

\begin{figure*}[t]
\centering
  \includegraphics[trim=0cm 0cm 0cm 0cm, clip=true, width=0.8\textwidth]{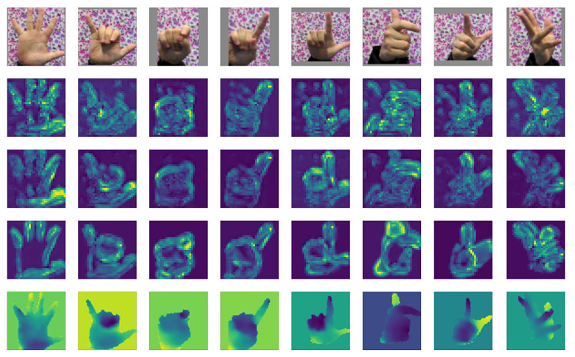}  
  \caption{\textbf{Feature activation maps}. \it (top row) input images, (row 2) activations of the RGB network trained on RGB only, (row 3) activations of the RGB network trained with additional depth data,  (row 4) activations of the depth network, and (row 5) depth images. During training, depth data helps the RGB network focus on the region of interest, reducing the influence of background regions.}
  
  \label{fig:feature_activations}
\end{figure*}

\begin{figure*}[t]
  \includegraphics[trim=0cm 8cm 0cm 0cm, clip=true, width=1.0\textwidth]{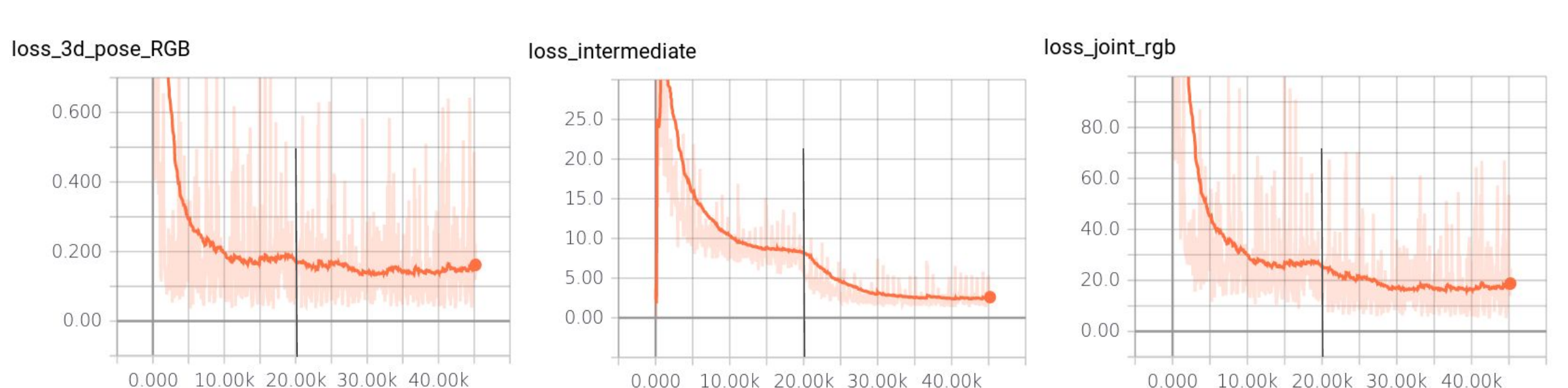}  
  \caption{\textbf{Loss function evolution on the stereo dataset~\cite{zhang20163d}.} \it The loss for 3D hand pose (left plot) of the RGB network on the test data converges at iteration 15,000, we continue training for another 5,000 iterations. From iteration 20,000, we fix the depth network parameters and connect mid-level features between the RGB and depth networks, and continue training by minimizing the joint loss (right plot) using RGB-D image pairs. The intermediate loss (middle plot) is used to suppress the difference between the mid-level feature between the RGB and depth networks. Loss for 3D hand pose of the RGB network, and the joint loss stop decreasing at around iteration 30k.}
  \label{fig:testing_loss}
\end{figure*}

\textbf{Feature activation maps:} 
To give more intuitions on the effectiveness of training using additional privileged information, we visualize the activations of the mid-level feature for the three networks. Feeding an RGB image into each network, we aggregate all the mid-level feature maps into feature map by taking the maximum across all feature maps (similar to the maxout operation~\cite{goodfellow2013maxout}). As shown in Figure~\ref{fig:feature_activations}, training with privileged information helps to select  more representative features, where the visualized activations are close to the foreground (the hand). 

\textbf{Loss function evolution:} We keep a record of the loss during our training on the Stereo dataset, see Figure~\ref{fig:testing_loss}.  The loss for 3D hand pose (left plot) of the RGB network on the test data converges at iteration 15,000, we continue training for another 5,000 iterations. From iteration 20,000, we fix the depth network parameters and connect mid-level features between the RGB and depth networks, and continue training by minimizing the joint loss (right plot) using RGB-D image pairs. The intermediate loss (middle plot) is used to suppress the difference between the mid-level feature between the RGB and depth networks. Loss for 3D hand pose of the RGB network, and the joint loss stop decreasing at around iteration 30,000.

\begin{figure*}[t]
  \includegraphics[trim=0cm 0cm 0cm 0cm, clip=true, width=1.0\textwidth]{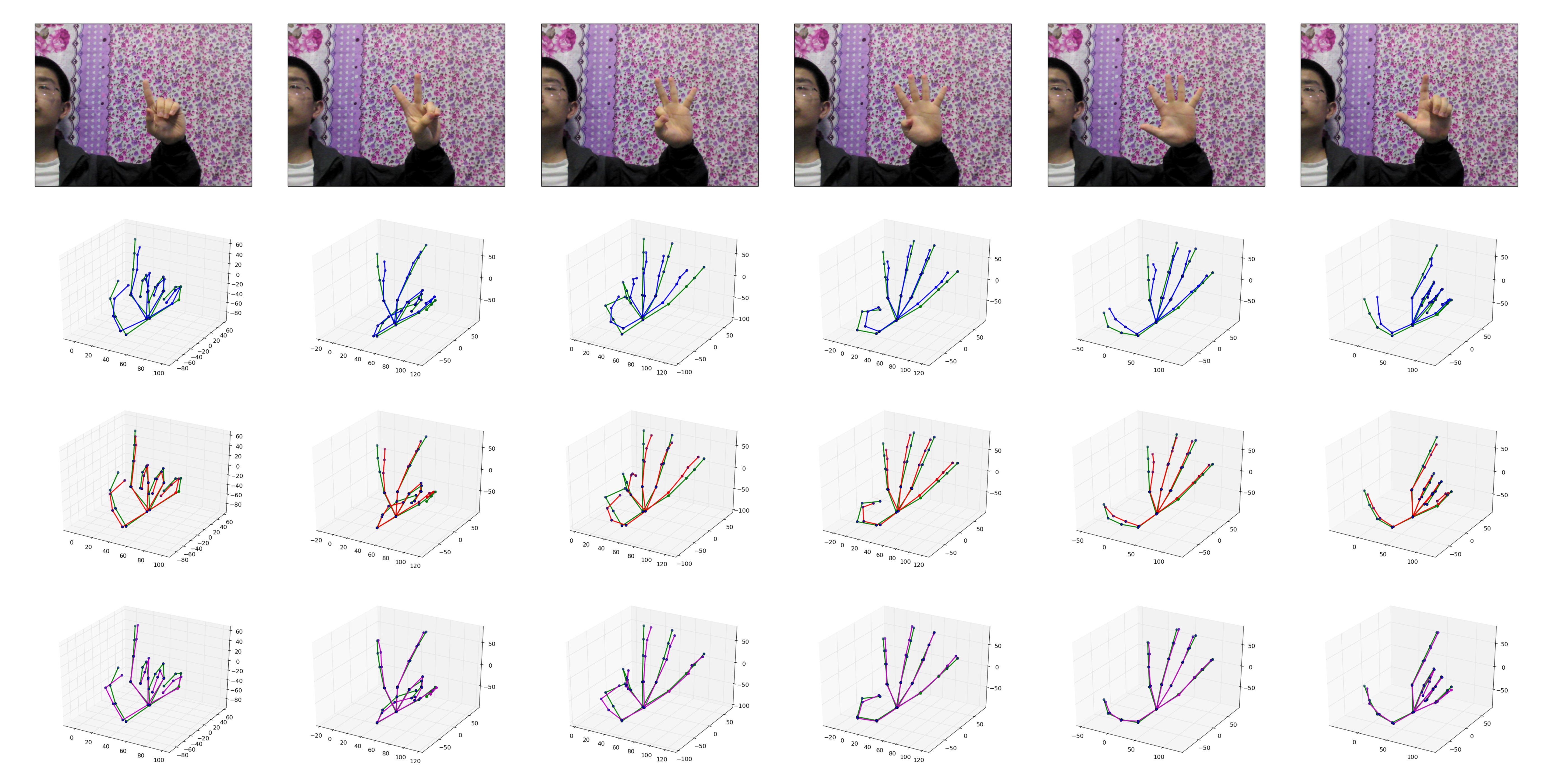}  
  \caption{\textbf{Qualitative 3D pose estimation results}.  \it Comparing the outputs of the RGB network (blue, second row), the RGB network with PI training (red, third row), and the Depth network (magenta, bottom row) with the ground truth 3D pose (green) on the Stereo dataset. Top row are the original images.}
  \label{fig:quali_stereo_3dpose}
\end{figure*}

\begin{figure*}[t]
  \includegraphics[trim=0cm 0cm 0cm 0cm, clip=true, width=1.0\textwidth]{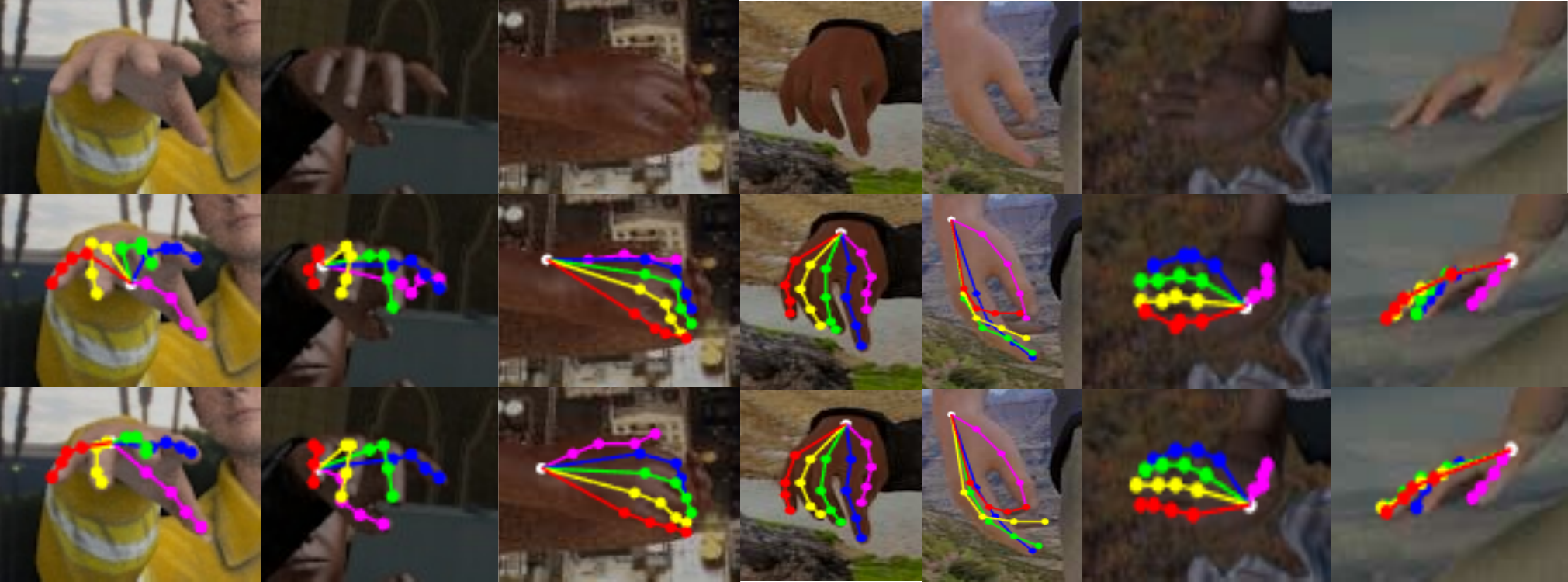}  
  \caption{\textbf{Qualitative 2D pose estimation results}.  \it Comparing the outputs of (middle) the RGB network and (bottom) the RGB network with PI training on the RHD dataset. Top row are the original images.}
  \label{fig:quali_rhd_2}
\end{figure*}

\begin{figure*}[t]
  \includegraphics[trim=0cm 0cm 0cm 0cm, clip=true, width=1.0\textwidth]{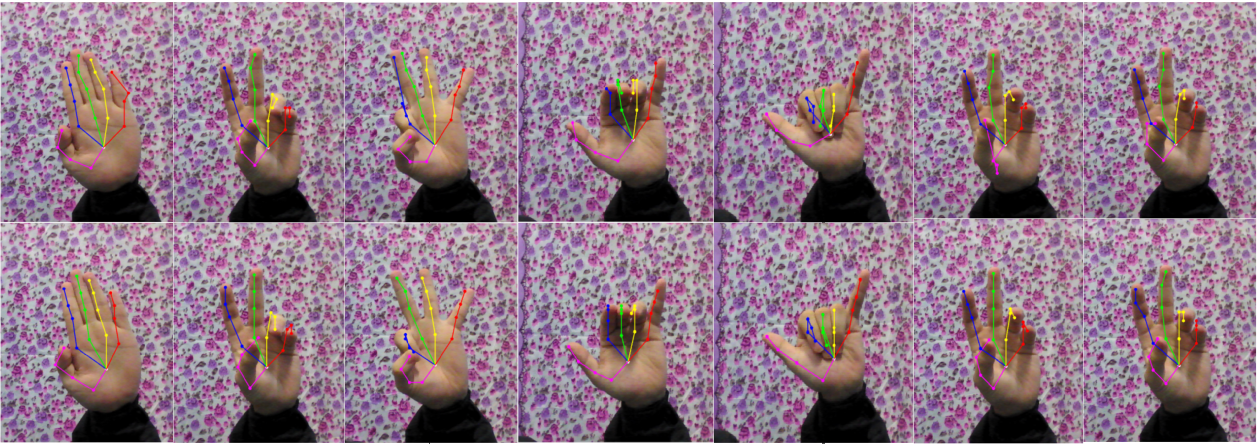}  
  \caption{\textbf{Qualitative 2D pose estimation results on Stereo dataset}.  \it Comparing the outputs of (top) the RGB network and (bottom) the RGB network with PI training.}
  \label{fig:quali_stereo_2}
\end{figure*}

\subsection{2D hand pose estimation from RGB}

In this section, we choose the base CNN model as CPM~\cite{wei2016convolutional}, which has shown great performance for 2D human pose estimation~\cite{wei2016convolutional}, and 2D hand pose estimation~\cite{zimmermann2017learning}.
Results are reported in Table~\ref{tab:RHDbaselines}, where `EPE' stands for the `average end point error' in pixels, where an end point is a hand joint. Qualitative examples are shown in Figure~\ref{fig:quali_rhd_2} and Figure~\ref{fig:quali_stereo_2}.
In this part of experiments, we treat the hand mask as privileged data, the CNN base model is CPM~\cite{simon2017hand}. The baseline is obtained by the normal training procedure, \ie, feeding the pre-processed hand image into CPM and obtaining the 2D hand pose by finding out the maximum location in each of the 21 heatmaps. For training with privileged information, we randomly select a certain proportion of RGB training data and use the hand masks, which are obtained by thresholding the depth images, in the \textit{Loss\_Mask} to suppress the background responses. As shown in Table~\ref{tab:RHDbaselines}, where 0.2 and 0.8 denotes the percentage of images when the \textit{Loss\_Mask} is used during the training for 2D hand pose estimation.

\textbf{Performance on hand-object interaction dataset:} 
In Figure~\ref{fig:sota_com} (bottom-right plot), we show a comparison in terms of 2D PCK (in pixels) on the  Dexter-Object~\cite{sridhar2016real} dataset. \textit{Z\&B\_Joint} denotes the method of Z\&B~\cite{zimmermann2017learning} trained on both RHD and Stereo datasets, which is better than \textit{Z\&B\_Stereo} (trained on Stereo) and \textit{Z\&B\_RHD} (trained on RHD). Our approach outperformed \textit{Z\&B\_Joint} even though we used less RGB training data.

\begin{table}
  \centering
  \small
  \resizebox{\columnwidth}{!}{
  \begin{tabular}{llllll}
  \toprule 
  \bf Method & \bf Testing & \bf Training & \bf EPE median & \bf EPE mean \\
  \midrule
  Z\&B~\cite{zimmermann2017learning} & RHD & RHD+Stereo  & 5.001  &  9.135 \\
  Baseline RGB & RHD & RHD  &  3.708  & 7.841 \\
  Baseline Depth & RHD & RHD   & 2.087 & 3.902 \\
  RGB + PI training & RHD & RHD   & 2.642 & 5.223 \\
  \hline    
   Z\&B~\cite{zimmermann2017learning} & Stereo & RHD+Stereo  &  5.522  &  5.013 \\ 
  Baseline RGB & Stereo & Stereo   & 5.250 & 6.533 \\
  Baseline Depth & Stereo & Stereo   & 4.775  & 5.883 \\
  RGB + PI training (0.2)& Stereo & Stereo   & 5.068   & 6.280 \\
  RGB + PI training (0.8)& Stereo & Stereo   & 4.515  & 5.801 \\

  \hline
  Z\&B~\cite{zimmermann2017learning} & Dexter-Object &   RHD+Stereo  &  13.684  &  25.160 \\ 
  Baseline RGB & Dexter-Object & RHD  &  13.360 & 18.278 \\
  RGB + PI training & Dexter-Object & RHD   & 11.809  & 14.593 \\  
 
   \bottomrule  
   \end{tabular}}
    \caption{\textbf{2D Hand Pose Accuracy}. \it Results when training on the RHD and Stereo datasets.
    } 
  \label{tab:RHDbaselines}
\vspace{-5mm}
\end{table}

\section{Conclusions}

In this paper, we proposed a framework for 3D hand pose estimation from RGB images, with the training stage aided with privileged information, {\it i.e.} depth data.
To the best of our knowledge, our method is the first to introduce the concept of using privileged information (depth images) to support training a RGB-based 3D hand pose estimator.   
We proposed three ways to use the privileged information: as external training data for a depth-based network branch, as paired depth data to transfer supervision from the depth-based network to the RGB-based network, and as a hand mask to suppress background activations in the RGB-based network.
Our training strategy can be easily embedded into existing pose estimation methods. As an illustration, we estimate 2D hand pose from an RGB image using a different CNN model. Results on 2D hand pose estimation, using our training strategy, are improved over state-of-the-art methods for 2D hand pose estimation from RGB input.  
During testing, when only RGB images are available, our model significantly outperforms the same model trained only using RGB images. 
This training strategy can be incorporated into existing models to boost the performance of hand pose estimation from an RGB image.
One limitation of our method is the difficulty of handling occlusion by objects, which can be addressed by systematically adding synthetic objects in the depth data (privileged information).

\textbf{Acknowledgement}: This work was supported
by Huawei Technologies.

{\small
\bibliographystyle{ieee}
\bibliography{egbib}
}

\end{document}